%% file: latex/acl_latex.tex
\documentclass[11pt]{article}

\usepackage[preprint]{acl}

\usepackage{times}
\usepackage{latexsym}

\usepackage[T1]{fontenc}

\usepackage[utf8]{inputenc}

\usepackage{microtype}

\usepackage{inconsolata}

\usepackage{graphicx}

\title{Incentivizing Temporal-Awareness in Egocentric Video Understanding Models}
\author{
  Zhiyang Xu\thanks{Work done while interning at Apple} \\
  Virginia Tech \\
  \texttt{zhiyangx@vt.edu}
  \And
  Tian Qin\footnotemark[1]\\
  Harvard University \\
  \texttt{tqin@g.harvard.edu}
  \And
  Bowen Jin\footnotemark[1]\\
  UIUC \\
  \texttt{bowenj4@illinois.edu}
  \AND
  Zhengfeng Lai \\
  UC Davis \\
  \texttt{lzhengfeng@ucdavis.edu}
  \And
  Meng Cao \\
  Apple \\
  \texttt{meng@apple.com}
  \And
  Lifu Huang\thanks{Corresponding authors} \\
  UC Davis \\
  \texttt{lfuhuang@ucdavis.edu}
  \And
  Peng Zhang\footnotemark[2] \\
  Apple \\
  \texttt{pzhang34@apple.com}
}

\input{latex/math}
\begin{document}
\maketitle
\input{latex/0_abstract}
\input{latex/1_intro}
\input{latex/2_related_work}
\input{latex/3_method}
\input{latex/4_experiment}

\input{latex/5_result}

\bibliography{custom}




\end{document}

%% file: latex/math.tex
\usepackage{graphicx}               
\usepackage{tabularx}               
\usepackage{booktabs}
\usepackage{amsmath}
\usepackage{stmaryrd}

\usepackage{soul}

\newcolumntype{C}{>{\centering\arraybackslash}X}
\usepackage{multirow}               
\usepackage{diagbox}                
\usepackage{hhline}                 
\usepackage{color,colortbl}         
\usepackage{amsmath}                
\usepackage{amssymb}                
\usepackage{mathtools}              
\usepackage{enumitem}               

\usepackage{subcaption}
\usepackage{caption}
\usepackage{booktabs}
\usepackage{extarrows}
\usepackage{makecell}
\usepackage{hyperref}
\usepackage{cleveref}
\usepackage{fdsymbol}
\usepackage{graphicx}
\usepackage{relsize}
\usepackage{wrapfig}
\usepackage{amsmath}
\usepackage[most]{tcolorbox}
\usepackage{booktabs}
\usepackage[table]{xcolor}

\usepackage{amssymb}
\usepackage{pifont}
\usepackage{xcolor}
\definecolor{CadetBlue}{RGB}{95,158,160}

\definecolor{TgpoPurple}{HTML}{BB85FF}
\definecolor{TgpoBlue}{HTML}{579EBB}

\definecolor{RoseQuartzBg}{HTML}{F7CAC9}
\definecolor{ForestGreen}{HTML}{228b22}
\definecolor{RoseQuartz}{HTML}{F5A798}
\definecolor{Serenity}{HTML}{92A8D1}
\definecolor{OrangeRed}{rgb}{1.0, 0.27, 0.0}
\definecolor{Red}{rgb}{1.0, 0.0, 0.0}
\definecolor{forestgreen}{rgb}{0.13, 0.55, 0.13}
\definecolor{Turquoise}{HTML}{0F4C81}
\definecolor{columbiablue}{rgb}{0.61, 0.87, 1.0}
\definecolor{Gray}{gray}{0.9}
\definecolor{lavenderPurple}{HTML}{7030A0}
\definecolor{methodblue}{HTML}{0070C0}

\usepackage{xparse}
\usepackage{footmisc}

\NewDocumentCommand{\ying}{ mO{} }{\textcolor{teal}{\textsuperscript{\textit{Ying}}\textsf{\textbf{\small[#1]}}}}

\NewDocumentCommand{\lifu}{ mO{} }{\textcolor{blue}{\textsuperscript{\textit{Lifu}}\textsf{\textbf{\small[#1]}}}}

\NewDocumentCommand{\zhiyang}{ mO{} }{\textcolor{Turquoise}{\textsuperscript{\textit{zhiyang}}\textsf{\textbf{\small[#1]}}}}

\definecolor{lightblue}{rgb}{0.0, 0.5, 1.0}

\definecolor{DarkCoffee}{HTML}{4B2E2B}
\definecolor{MediumCoffee}{HTML}{6F4E37}
\definecolor{LightCoffee}{HTML}{A67B5B}
\definecolor{Golden}{HTML}{D4A017}
\newcommand{\archcolor}[1]{{\textcolor{DarkCoffee}{L}\textcolor{DarkCoffee}{a}\textcolor{MediumCoffee}{T}\textcolor{MediumCoffee!80}{t}\textcolor{MediumCoffee}{E}-\textcolor{Golden}{F}\textcolor{Golden!90}{l}\textcolor{Golden!80}{o}\textcolor{Golden!70}{w}}}

\newcommand{\archlong}[1]{{Layerwise Timestep-Expert Flow-based Transformer}} %

\newcommand{\trpolong}[1]{Temporal Global Policy Optimization}
\newcommand{\tspolong}[1]{Temporal Batch-Sequence Policy Optimization}

\newcommand{\trpo}[1]{\textsc{TGPO}}
\newcommand{\tspo}[1]{{T-BSPO}}

\definecolor{Espresso}{HTML}{4B2E2B}
\definecolor{Mocha}{HTML}{6F4E37}
\definecolor{Latte}{HTML}{A67B5B}
\definecolor{Cappuccino}{HTML}{D2B48C}
\definecolor{Caramel}{HTML}{C08051}
\definecolor{Cream}{HTML}{F5F5DC}

%% file: latex/0_abstract.tex
\begin{abstract}
Multimodal large language models (MLLMs) have recently shown strong performance in visual understanding, yet they often lack temporal awareness, particularly in egocentric settings where reasoning depends on the correct ordering and evolution of events. This deficiency stems in part from training objectives that fail to explicitly reward temporal reasoning and instead rely on frame-level spatial shortcuts. To address this limitation, we propose \trpolong{} (\trpo{}), a reinforcement learning with verifiable rewards (RLVR) algorithm designed to incentivize temporal awareness in MLLMs. \trpo{} contrasts model outputs generated from temporally ordered versus shuffled video frames to derive calibrated, globally normalized reward signals that explicitly favor temporally coherent reasoning. Integrated with GRPO and GSPO, \trpo{} supports cold-start RL training and effectively suppresses spatial shortcut behaviors learned by existing MLLMs. Experiments across five egocentric video benchmarks demonstrate that \trpo{} consistently improves temporal grounding and causal coherence, outperforming prior RL-based video reasoning approaches. Our results suggest that \trpo{} offers a simple and scalable pathway toward temporally robust MLLMs for egocentric video understanding.
\end{abstract}

%% file: latex/1_intro.tex
\section{Introduction}
\label{sec:intro}

Egocentric video understanding focuses on videos captured from a first-person perspective, where the camera is worn by an agent and closely follows the agent’s actions, attention, and interactions with the environment~\cite{ego4d,egoplan,egoschema}. Compared to third-person videos with relatively stable viewpoints, egocentric videos exhibit rapid viewpoint changes, severe partial observability, and strong causal dependencies between past and future frames~\cite{ego4d}. These characteristics substantially amplify the difficulty of temporal reasoning.

\begin{figure}[t]
  \centering
  \includegraphics[width=\linewidth]{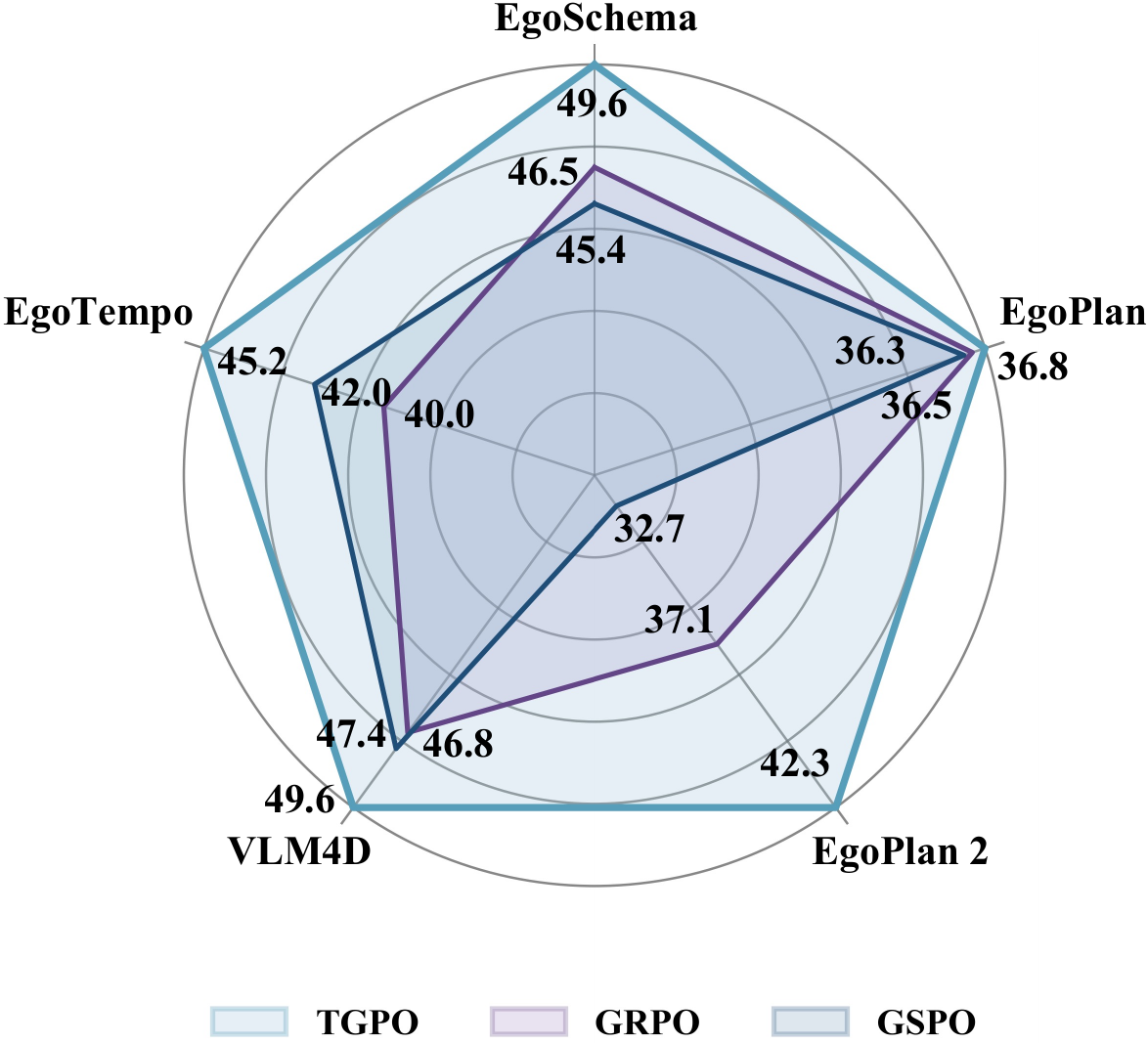}
  \caption{Performance comparison of \trpolong{} (\trpo{}), GRPO, and GSPO on five egocentric benchmarks.}
  \label{fig:radar}
\end{figure}

Core egocentric tasks such as action recognition~\cite{egoschema}, temporal grounding~\cite{egoApple}, and intention prediction~\cite{egoplan,egoplan2} inherently require models to maintain temporal coherence and reason over long-term temporal dependencies~\cite{egoschema,egoplan,egotempo}. Most existing egocentric benchmarks are formulated as video-question answering tasks~\cite{qaego4d}, where the model must answer a question based on information distributed across the video rather than from any single frame. Despite recent progress in multimodal large language models (MLLMs)~\cite{egovideo,exo2ego,egoApple}, current approaches frequently fail in such settings, exhibiting hallucinated events, temporally inconsistent reasoning, and degraded performance.

We identify three fundamental challenges that limit the temporal reasoning capability of existing MLLMs in egocentric video understanding. \textbf{First}, prevailing post-training paradigms, including supervised fine-tuning and reinforcement learning methods such as Group Relative Policy Optimization (GRPO)~\cite{grpo}, optimize answer correctness but do not explicitly incentivize temporal consistency or causal understanding across frames~\cite{videoR1}. As a result, models may produce correct answers without genuinely reasoning over temporal structure. \textbf{Second}, many training datasets contain questions that are answerable from a single frame, allowing models to rely on spatial shortcuts instead of learning temporal dynamics~\cite{breakdownvideobench}. This encourages degenerate behaviors that treat videos as unordered collections of images. \textbf{Third}, existing benchmarks rarely provide high-quality temporal reasoning chains. Most datasets offer only video–answer pairs, while temporally grounded, long-horizon rationales require costly human annotation and are therefore scarce~\cite{egovlm}. Together, these limitations lead to models that struggle with causal dependencies and generate temporally incoherent or weakly supported explanations.

To address these challenges, we propose \textbf{\trpolong{} (\trpo{})}, a novel reinforcement learning with verifiable rewards (RLVR) algorithm that explicitly incentivizes temporal awareness in MLLMs. At a high level, \trpo{} introduces temporal calibration into the reward objective. During training, the model first consumes temporally ordered video frames and generates diverse responses via temperature sampling. In parallel, for the same video–question pairs, we remove temporal information by randomly shuffling the video frames and generate a response using greedy decoding. The reward obtained from this shuffled-video generation serves as a baseline representing performance without temporal cues. We then subtract this baseline from the reward of each sampled response, yielding a temporally calibrated reward signal. This formulation explicitly rewards improvements that arise from leveraging correct temporal ordering. The calibrated rewards are further normalized across both training instances and sample groups, which we refer to as \emph{global normalization}, in contrast to conventional group normalization, enabling more stable and effective policy updates.

In our experiments, we integrate \trpo{} into two widely used policy optimization frameworks, GRPO and Group Sequence Policy Optimization (GSPO) \cite{gspo}, and apply them to train Qwen2.5-VL models~\cite{Qwen2.5vl}. Following the DeepSeek-R1-Zero paradigm~\cite{deepseekR1}, we adopt a cold-start training regime, directly optimizing the MLLMs with \trpo{} without any supervised fine-tuning stage. We evaluate the resulting models on five egocentric video benchmarks that require strong temporal awareness and long-horizon reasoning to achieve high performance. Across all benchmarks, Qwen2.5-VL trained with \trpo{} consistently outperforms prior reinforcement learning–based approaches, demonstrating substantial gains on temporal reasoning tasks.

In summary, our contributions are threefold:
\begin{itemize}
\item We introduce \trpo{}, a temporally calibrated RLVR framework that explicitly incentivizes temporal awareness in multimodal large language models.
\item We mitigate the influence of temporally ambiguous or low-quality training signals, preventing models from learning spurious spatial shortcuts.
\item We demonstrate that cold-start reinforcement learning alone is sufficient for strong temporal reasoning performance, eliminating the need for human-labeled reasoning chains and simplifying the training pipeline.
\end{itemize}

%% file: latex/2_related_work.tex
\section{Related Works}

\subsection{Policy Optimization Algorithms}
A wide range of policy-optimization algorithms has been explored for post-training large language models (LLMs). Proximal Policy Optimization (PPO)~\cite{ppo} introduces a clipped surrogate objective that stabilizes policy updates and has become a foundational baseline in reinforcement learning from human feedback (RLHF)~\cite{rlhf}. Direct Preference Optimization (DPO)~\cite{dpo} instead learns directly from preference pairs without training a reward model, simplifying alignment by optimizing likelihood ratios between preferred and disfavored responses. More recent critic-free approaches include ReMax~\cite{remax}, which adapts REINFORCE with a greedy baseline for simplicity, and RLOO~\cite{RLOO}, which reduces variance by subtracting a leave-one-out baseline computed from other sampled responses. REINFORCE++~\cite{reinforce++} further advances this line of work by using a batch-normalized reward baseline for advantage estimation, avoiding an explicit critic while improving robustness and generalization across reward models and long chain-of-thought settings. Group-based algorithms such as GRPO~\cite{grpo} normalize rewards within each sample group to estimate relative advantages without a value network, while GSPO~\cite{gspo} extends this idea by performing optimization at the sequence level to better match sequence-level rewards. In contrast, Single-stream Policy Optimization (SPO)~\cite{spo} revisits policy-gradient learning from a non-grouped perspective, replacing per-group baselines with a persistent KL-adaptive value tracker and globally normalized advantages, enabling stable, low-variance learning signals and significantly improved scalability.

\subsection{Ego-Centric Video Understanding}

Ego4D~\cite{ego4d} introduced a large-scale egocentric video dataset with 3,670 hours of multimodal first-person recordings and a comprehensive benchmark suite covering episodic memory, interaction understanding, and future activity forecasting, establishing a foundational resource for egocentric perception research. EgoVLPv2~\cite{egovlpv2} advanced egocentric video–language pretraining by integrating cross-modal fusion directly into the backbone networks, enabling more unified representations and reducing downstream fine-tuning costs. GroundNLQ~\cite{GroundNLQ} proposed a multi-scale multimodal grounding framework for long egocentric videos and achieved state-of-the-art performance in the Ego4D Natural Language Queries Challenge through specialized egocentric feature extraction. EMQA~\cite{qaego4d} introduced episodic memory–based video question answering, constraining models to maintain constant-sized memory representations and releasing the large-scale QAEGO4D dataset to study long-horizon egocentric reasoning. EgoVideo \cite{egovideo} presented an egocentric foundation model tailored for Ego4D and EPIC-Kitchens challenges, demonstrating strong generalization across diverse egocentric tasks, including moment retrieval and action anticipation. MM-EGO~\cite{mmego} explored building egocentric multimodal LLMs by generating 7M QA pairs from Ego4D, proposing a memory pointer prompting mechanism to improve long-video comprehension and de-biased evaluation. EgoLife~\cite{egolife} introduced a life-oriented egocentric dataset and QA benchmark spanning daily activities over extended time horizons, along with an integrated assistant system combining multimodal modeling and retrieval for long-context reasoning. EgoVLM~\cite{egovlm} applied Group Relative Policy Optimization to directly align vision–language models with egocentric reasoning behaviors, demonstrating substantial gains over general-purpose VLMs and introducing a keyframe-based reward for temporal grounding. Ego-R1~\cite{egor1} proposed a Chain-of-Tool-Thought framework with an RL-trained agent to reason over ultra-long egocentric videos, enabling modular tool invocation for temporal retrieval and multimodal understanding across week-long time spans. EgoVITA~\cite{egovita} introduced a reinforcement learning framework that alternates between egocentric planning and exocentric verification, improving causal and visually grounded reasoning for first-person video understanding. Exo2Ego~\cite{exo2ego} leveraged large-scale synchronized ego–exo video–text data to transfer exocentric knowledge into egocentric domains through progressive mapping, significantly improving egocentric performance while highlighting limitations of existing MLLMs.

%% file: latex/3_method.tex
\begin{figure*}[h]
  \centering
  \includegraphics[width=0.9\textwidth]{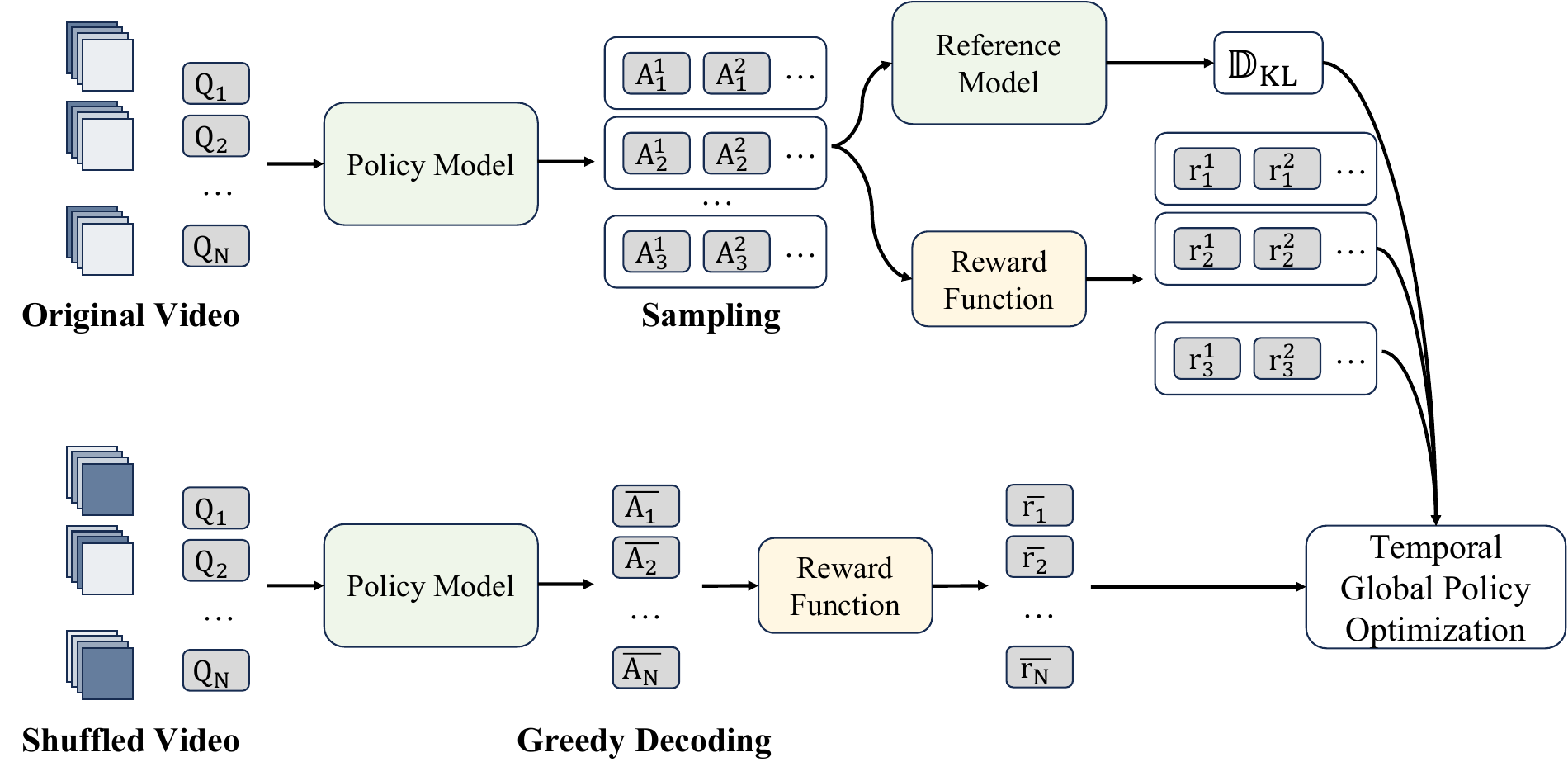}
  \caption{An overview of our proposed \trpo{}. In the upper half, the original video is passed into the policy model, and a group of responses are sampled. Rewards are assigned to each response in the group. In the bottom half, the policy model takes in a shuffled video and performs greedy decoding to generate responses. The reward model assigns scores to responses without temporal information.}
  \label{fig:task_split}
\end{figure*}

\section{Method}

\subsection{Background on GRPO}
In the GRPO framework~\cite{grpo}, given an input prompt $s$ and a video clip $c$ as context, the MLLM samples a group of $|G|$ candidate responses $\{y_1,\ldots,y_{|G|}\}$. A reward function $r(\cdot)$ assigns a scalar score to each response, producing \\ $\{r(y_1),\ldots,r(y_{|G|})\}$. GRPO optimizes the policy by maximizing a group-normalized advantage estimator:
\begin{align*}
\hat{A}(s,c;\theta) &= \frac{1}{|G|}\sum_{i=1}^{|G|}\frac{1}{|y_i|}\sum_{t=1}^{|y_i|}
\frac{\pi_\theta(y_{i,t}\mid s,c)}{\pi_{\theta_{\text{old}}}(y_{i,t}\mid s,c)}\\&
\cdot
\frac{r(y_i)-\mu_G}{\sigma_G},
\end{align*}
where $\pi_\theta(y_{i,t}\mid s,c)$ denotes the log-probability of generating token $y_{i,t}$ under the current parameters $\theta$, and $\pi_{\theta_{\text{old}}}$ corresponds to a recently updated policy used to form the importance ratio. Here $\mu_G=\mathrm{mean}(\{r(y_i)\}_{i=1}^{|G|})$ and $\sigma_G=\mathrm{std}(\{r(y_i)\}_{i=1}^{|G|})$ are computed within the sampled group for the given $(s,c)$.

To stabilize training and prevent excessive drift from a reference policy $\pi_{\text{ref}}$, GRPO includes a KL regularizer. The resulting objective is
\begin{equation*}
\max_{\theta}\ \mathbb{E}_{(s,c)\sim\mathcal{D}}
\left[
\hat{A}(s,c;\theta)\;-\;\beta\,D_{\mathrm{KL}}\!\left(\pi_\theta\,\|\,\pi_{\text{ref}}\right)
\right],
\end{equation*}
where $\beta$ controls the strength of the regularization.

\subsection{Greedy Baseline without Temporal Information}
We introduce a temporally calibrated reward for video understanding that explicitly encourages MLLMs to exploit temporal cues, rather than relying on static per-frame shortcuts.

Given a prompt $s$ and a video $c$, we first sample a response from the current policy, $y\sim \pi_\theta(\cdot\mid s,c)$. We then construct a temporal-calibrated baseline by shuffling the video frames and generate a response with greedy decoding:
\[
\hat{y}\sim \pi_\theta(\cdot\mid s,\mathrm{shuffle}(c)),
\]
We define the temporally calibrated reward as
\begin{equation*}
r_{\mathrm{T}}(y) \;=\; r(y)\;-\;r(\hat{y}).
\end{equation*}
Compared to $r(y)$, the temporally calibrated reward takes into consideration whether temporal reasoning is used in the model's generation. 
Specifically, when $r_{\mathrm{T}}(y)>0$, the model performs better with the temporally coherent video than with shuffled frames, indicating that it leverages temporal dependencies; we therefore assign a positive training signal. Conversely, $r_{\mathrm{T}}(y)\le 0$ suggests that the prediction does not benefit from temporal information (e.g., the model treats the input as a set of independent frames), and the resulting non-positive signal discourages such shortcut behavior.

\subsection{\trpolong{} (\trpo{})}
We next show that \trpo{} can be incorporated into existing policy-optimization objectives. We present two variants based on GRPO and GSPO.

\paragraph{Integration with GRPO.}
Standard GRPO normalizes rewards within each group of size $|G|$ for each prompt $s_j$ in a mini-batch $B$. In contrast, \trpo{} performs normalization across all group samples in the mini-batch, which prevents low-variance groups (often corresponding to temporally insensitive instances) from being artificially amplified by within-group normalization.

Concretely, let $|B|$ be the batch size and $|G|$ the group size. For instance $j$, if the rewards of the group samples $\{r(y_{j,1}),\ldots,r(y_{j,|G|})\}$ are close to the baseline reward $r(\hat{y}_j)$, then the calibrated rewards $r_{\mathrm{T}}(y_{j,i})$ are near zero, indicating that temporal information is largely unnecessary for answering $s_j$. With only in-group normalization, such low-variance groups can still yield large normalized advantages and thus contribute disproportionately to training, encouraging temporally insensitive behavior. Global normalization mitigates this issue: if other instances in the same mini-batch achieve large $r_{\mathrm{T}}(\cdot)$, the batch-level mean increases and the normalized advantage for temporally insensitive instances becomes small or negative, reducing their influence.
The resulting \trpo{} (GRPO) advantage estimator is
\begin{align*}
\hat{A}(B;\theta) &= \frac{1}{|B|}\sum_{j=1}^{|B|}\frac{1}{|G|}\sum_{i=1}^{|G|}
\frac{1}{|y_{j,i}|}\\&\sum_{t=1}^{|y_{j,i}|}
\frac{\pi_\theta(y_{j,i,t}\mid s_j,c_j)}{\pi_{\theta_{\text{old}}}(y_{j,i,t}\mid s_j,c_j)}
\cdot 
\frac{r_{\mathrm{T}}(y_{j,i})-\mu_B}{\sigma_B},
\end{align*}
where $\mu_B=\mathrm{mean}(\{r_{\mathrm{T}}(y_{j,i})\}_{j=1,i=1}^{|B|,|G|})$ and $\sigma_B=\mathrm{std}(\{r_{\mathrm{T}}(y_{j,i})\}_{j=1,i=1}^{|B|,|G|})$ are computed over all $|B|\!\times\!|G|$ samples in the mini-batch.

\paragraph{Integration with GSPO.}
GSPO~\cite{gspo} replaces token-level importance ratios with a sequence-level likelihood ratio, which empirically improves stability while achieving performance comparable to GRPO. Under GSPO, we define the sequence-level importance ratio as
\[
\rho_{j,i}(\theta) \;=\;
\exp\!\left(
\frac{1}{|y_{j,i}|}\sum_{t=1}^{|y_{j,i}|}
\log\frac{\pi_\theta(y_{j,i,t}\mid s_j,c_j)}{\pi_{\theta_{\text{old}}}(y_{j,i,t}\mid s_j,c_j)}
\right),
\]
then the \trpo{} (GSPO) advantage estimator becomes
\begin{align*}
\hat{A}(B;\theta) = \frac{1}{|B|}\sum_{j=1}^{|B|}\frac{1}{|G|}\sum_{i=1}^{|G|}
\rho_{j,i}(\theta)
\cdot
\frac{r_{\mathrm{T}}(y_{j,i})-\mu_B}{\sigma_B},
\end{align*}

\subsection{Prompt Engineering.}
During rollout, we prompt the model to generate not only the final answer but also a reasoning trace, formatted in a required structure. Following prior work~\cite{deepseekR1}, we design the input prompt to encourage the model to produce an answer accompanied by step-by-step reasoning. The exact prompts are provided below.

\begin{tcolorbox}[
  colback=gray!10,
  colframe=gray!60,
  boxrule=0.5pt,
  arc=2mm,
  left=3mm,right=3mm,top=2mm,bottom=2mm
]
I provide you with a question about the given video.

Provide a detailed thinking process within \texttt{<think> </think>} tags and then output the answer within the \texttt{<answer> </answer>} tags.
Within \texttt{<think>} tags, provide a detailed, step-by-step reasoning process. First, describe in detail what happens in the video that is relevant to the question. Then, explain how you arrive at your answer by referencing specific evidence or events from the video. Make sure your reasoning is clear, logical, and closely tied to what is shown in the video.
Within \texttt{<answer>} tags, clearly state your chosen answer, ensuring you select it from the provided options.

The question is: [EVENT]
\end{tcolorbox}

\subsection{Reward Modeling}
We employ a composite reward that evaluates both \emph{answer correctness} and \emph{output compliance}. Since our datasets use multiple-choice video question answering (VQA), each model response is expected to contain (i) a reasoning segment enclosed by \texttt{<think>...</think>} and (ii) a final choice enclosed by \texttt{<answer>...</answer>}. The overall reward is computed per sampled response and used for policy optimization.

\paragraph{Accuracy Reward.}
Let $\bar{\alpha}$ denote the ground-truth option for a given question, and let $a$ be the option extracted from the model output (i.e., the content inside \texttt{<answer>...</answer>} after normalization). We define a binary accuracy reward:
\[
r_{\text{Accu}}(a) = 
\begin{cases}
0, & \text{if } a \neq \bar{\alpha} \\
1, & \text{if } a = \bar{\alpha}
\end{cases}
\]

\paragraph{Format Reward.}
In addition to correctness, we compute reward adherence to the required response structure. 
Specifically, we assign:
\[
r_{\text{Form}}(a) = 
\begin{cases}
0, & \text{not follow the required format} \\
1, & \text{follow the required format}
\end{cases}
\]
A response is considered well-formatted if it contains exactly one \texttt{<think>}...\texttt{</think>} block and one \texttt{<answer>}...\texttt{</answer>} block; 
and the \texttt{<answer>} block is non-empty and contains a valid option from the provided candidate set (after normalization). 

\paragraph{Combined Reward.}
We combine correctness and formatting as 
\[r(a)=r_{\text{Accu}}(a)+\lambda r_{\text{Form}}(a),\] 
where $\lambda$ controls the strength of the format signal.

%% file: latex/4_experiment.tex
\section{Experiments}
\subsection{Implementation Details}

\paragraph{Base Model and RL Framework.}
We adopt Qwen2.5-VL-3B~\cite{Qwen2.5vl} as our base model due to its strong performance on video understanding tasks and its demonstrated potential for RL–based optimization~\cite{videoR1,timer1,videochatr1}. All experiments are conducted using this backbone.

We train the model using \textsc{verl}~\cite{verl}, a flexible and efficient reinforcement learning framework designed for large-scale model training. At the time of our experiments, \textsc{verl} did not natively support video-based RL training. We therefore extend its implementation to enable video input processing and video understanding with \textsc{vLLM}~\cite{vllm}. In addition, we implement the GSPO baseline following the original formulation in prior work \cite{gspo}.

Unless otherwise specified, we use the same training hyper-parameters across all experiments: format-reward weight $\lambda = 0.1$, learning rate $1\times10^{-6}$ with a constant scheduler, KL regularization coefficient $1\times10^{-4}$, weight decay $0.01$, number of rollouts $8$, sampling temperature $1.0$, micro-batch size of $4$ per GPU, and mini-batch size $64$. Training is performed on $8$ nodes, each equipped with $8$ NVIDIA A100 GPUs with $40$GB memory.

\paragraph{Training Data.}
We use EgoIT99K~\cite{EgoIT99K} as the training dataset. Since our method and all baselines (Section~\ref{sec:baselines}) rely on verifiable rewards, we restrict training to the subsets of EgoIT99K that contain multiple-choice and yes/no questions.

For each video, we uniformly sample $32$ frames. Increasing the number of frames did not yield noticeable improvements in our preliminary experiments. Due to the lack of large-scale, high-quality supervised finetuning data for egocentric video understanding, we follow the cold-start training paradigm of DeepSeek-R1-Zero~\cite{deepseekR1}, training the model directly with reinforcement learning without a supervised pretraining stage.

\subsection{Evaluation}

\paragraph{Evaluation Datasets and Metrics.}
We evaluate all models on five egocentric video question-answering benchmarks designed to assess temporal understanding. All evaluation datasets consist of multiple-choice questions.
Prior work~\cite{breakdownvideobench} has shown that the full EgoSchema~\cite{egoschema} benchmark contains instances that can often be solved using language priors or static spatial cues from a single frame. To more faithfully evaluate temporal reasoning, we adopt the temporal and \emph{others} splits filtered by~\cite{breakdownvideobench}.
EgoPlan Bench~\cite{egoplan} and EgoPlan2 Bench~\cite{egoplan2} evaluate planning and anticipation capabilities, requiring models to observe ongoing actions in egocentric videos and predict the next step. EgoPlan2 further increases diversity and realism by grounding tasks in more complex real-world scenarios.
VLM4D~\cite{vlm4d} consists of carefully curated egocentric video–question pairs that emphasize translational and rotational motion, perspective awareness, and motion continuity. Finally, EgoTempo~\cite{egotempo} focuses on holistic temporal understanding, where correct answers cannot be inferred from a single frame or common-sense reasoning alone. 
Following~\cite{exo2ego}, we modify the answer of instances in  EgoTempo~\cite{egotempo} from direct answer generation to multiple-choice QA for a more stable and reliable evaluation. To construct high-quality distractors, we leverage a state-of-the-art vision-language model (VLM). Specifically, we provide the Gemini 2.5 model with the video instances and prompt it to generate plausible incorrect options based on the visual context and temporal proximity to the ground-truth answer.

\begin{table*}[h]
\centering
\resizebox{0.8\textwidth}{!}{
\begin{tabular}{l|ccccc}
\toprule
\rowcolor{gray!15}
\textbf{Model} 
& \textbf{EgoSchema$\uparrow$} 
& \textbf{EgoPlan$\uparrow$} 
& \textbf{EgoPlan 2$\uparrow$} 
& \textbf{VLM4D$\uparrow$} 
& \textbf{EgoTempo$\uparrow$} \\
\midrule
\multicolumn{6}{l}{\small\textbf{Qwen2.5-VL 3B}} \\
\midrule
+ CoT  & 20.4 & 22.2 & 23.5 & 37.1 & 31.6 \\
+ GRPO & 46.5 & 36.5 & 37.1 & 46.8 & 40.0 \\
+ GSPO & 45.4 & 36.3 & 32.7 & 47.4 & 42.0 \\
\midrule
\midrule
\rowcolor{TgpoPurple!20}
+ \trpo{} (GRPO) & 49.6 & 36.8 & 42.3 & 49.6 & 45.2 \\
\rowcolor{TgpoBlue!20}
+ \trpo{} (GSPO) & 49.7 & 36.7 & 41.1 & 48.6 & 42.6 \\
\bottomrule
\end{tabular}
}
\caption{Performance comparison of our method \trpo{} with two popular RL-based optimization methods and chain-of-thought (CoT) reasoning across egocentric benchmarks.}
\label{tab:rl_results}
\end{table*}

\subsection{Baselines}
\label{sec:baselines}
Our main contribution is the proposed RL algorithm \trpo{} for better temporal understanding and reasoning. Hence, our main comparison is focusing on comparing our approach against two RLVR baselines, including GRPO~\cite{grpo} and GSPO~\cite{gspo}. GRPO has been widely adopt for training strong video reasoning models including Video-R1~\cite{videoR1}, Time-R1~\cite{timer1}, and Video-Chat-R1,~\cite{videochatr1}. In our implementation, we keep everything for GRPO and GSPO, including training data and hyperparameters, the same as our methods for a fair comparison.

In addition, we benchmark against strong MLLMs including proprietary models: Gemini-1.5-Pro~\cite{gemini1.5}, Gemini-2.5-Pro~\cite{gemini2.5}, and Claude-Sonnet-4~\footnote{\url{https://www.anthropic.com/claude/sonnet}}; open-source models: Qwen2.5-VL~\cite{Qwen2.5vl}, LLaVA-Video~\cite{llavavideo}, LLaVA-NeXT-Video~\cite{llavanextvideo}, Video-LLaMA-2~\cite{videollama2}, Llava-OneVision~\cite{llavaonevision}, and InternVideo2~\cite{internvideo2}; and the egocentric-specialized model EgoVLM~\cite{egovlm}, which has been trained with GRPO.

%% file: latex/5_result.tex
\section{Results}

\subsection{Comparison with RLVR Methods}

\paragraph{Overall Performance.} As shown in Table \ref{tab:rl_results}, \trpo{} consistently outperforms existing RL-based approaches across all benchmarks. While standard CoT prompting performs poorly on temporally demanding tasks, GRPO and GSPO yield substantial improvements, particularly on EgoSchema and EgoPlan 2. However, these methods do not explicitly enforce temporal consistency, limiting their effectiveness on benchmarks requiring long-horizon temporal reasoning. Integrating \trpo{} on top of GRPO or GSPO leads to further and consistent gains. In particular, Qwen2.5-VL 3B + TGPO (GRPO) achieves the strongest overall performance, reaching 49.6 on EgoSchema, 36.8 on EgoPlan, 42.3 on EgoPlan 2, 49.6 on VLM4D, and 45.2 on EgoTempo. The largest improvements are observed on EgoSchema and EgoPlan 2, indicating that temporally calibrated rewards effectively discourage single-frame shortcuts and promote reasoning over event order and temporal dependencies. Overall, these results demonstrate that \trpo{} substantially enhances temporal reasoning in MLLMs under a fully reinforcement-learning-based training regime, outperforming existing RL methods.

\begin{figure*}[h!]
  \centering
  \includegraphics[width=\textwidth]{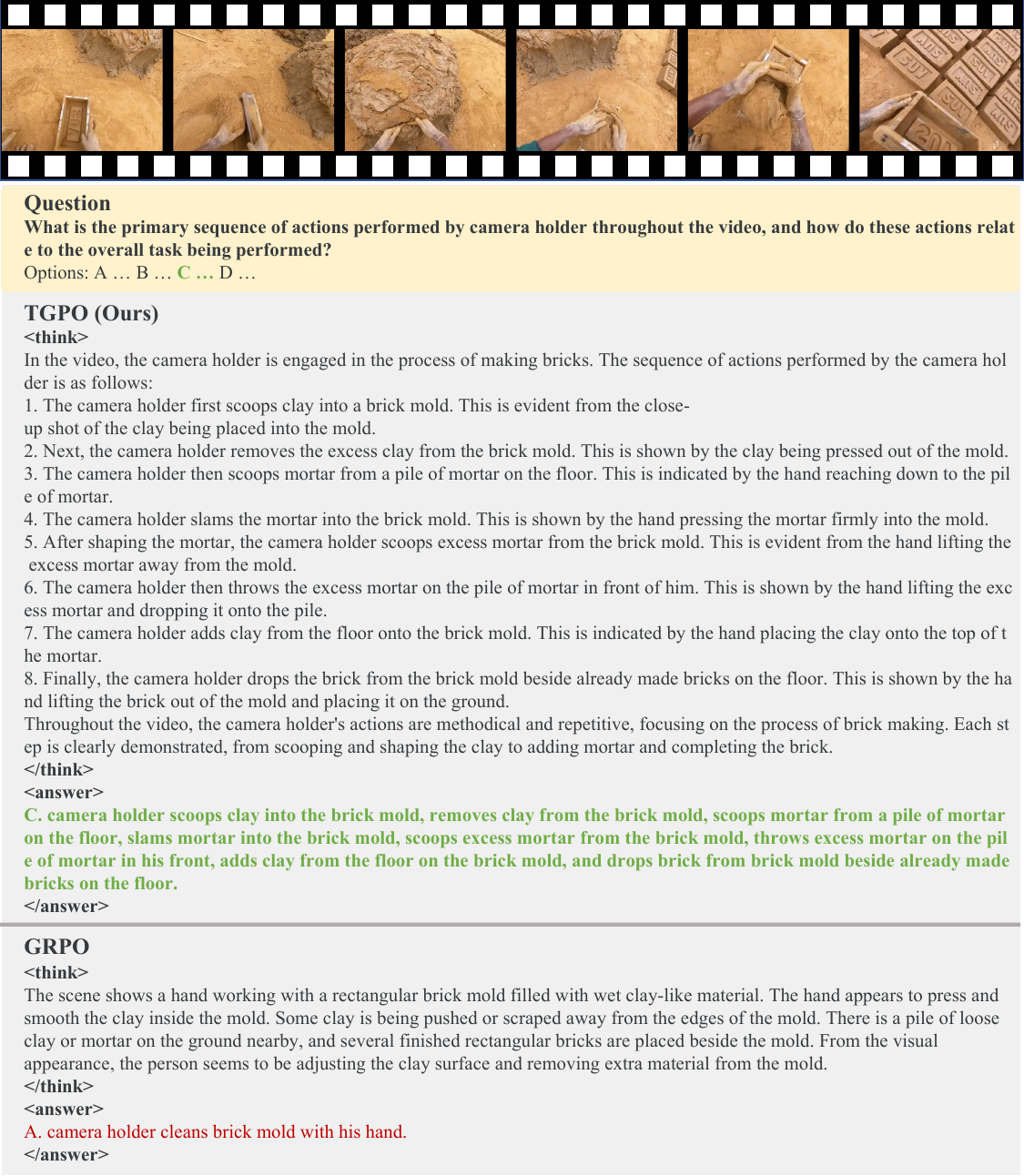}
  \caption{Qualitative comparison between \trpo{} and GRPO on the EgoSchema Benchmark.}
  \label{fig:qualitative}
\end{figure*}

We provide a qualitative comparison between \trpo{} (GRPO) and the GRPO baseline in Figure~\ref{fig:qualitative}. The example requires understanding the temporal order of a multi-step brick-making process. As shown in the reasoning chain, \trpo{} correctly tracks the chronological sequence of actions and produces a temporally coherent explanation. In contrast, GRPO mainly relies on salient frame-level spatial cues (e.g., a hand interacting with a mold) and collapses the procedure into an incorrect single-step interpretation (e.g., “cleans clay out from the mold”), failing to capture the temporal relations across frames. This highlights that \trpo{} reduces spatial shortcuts and encourages temporally grounded reasoning.

\paragraph{Training Dynamics.}

\begin{table}[t]
\centering
\setlength{\tabcolsep}{2pt}  
\resizebox{\linewidth}{!}{
\begin{tabular}{l c c c c}
\toprule
\rowcolor{gray!15}
\textbf{Method} & \textbf{VLM4D} & \textbf{EgoPlan} & \textbf{EgoPlan2} & \textbf{EgoSchema} \\
\midrule

\multicolumn{5}{l}{\small\textbf{Baseline}} \\
GRPO      & 1265.13 & 898.25 & 784.04 & 1159.03 \\
GSPO      & 1305.99 & 693.21 & 552.01 & 1118.76 \\
\midrule\midrule

\multicolumn{5}{l}{\small\textbf{Ours}} \\
\rowcolor{TgpoPurple!20}
TGPO (GRPO) & \textbf{1339.49} & \textbf{923.93} & \textbf{786.97} & \textbf{1306.32} \\
\rowcolor{TgpoBlue!20}
TGPO (GSPO) & 1333.75 & 905.82 & 544.95 & 1144.84 \\
\bottomrule
\end{tabular}
}
\caption{Area Under the Curve (AUC) of reward over the first 3000 training steps for different optimization methods across datasets. A higher AUC reflects faster reward improvement and improved training stability.}
\label{tab:auc_across_datasets}
\end{table}

\begin{figure*}[h]
  \centering
  \includegraphics[width=\textwidth]{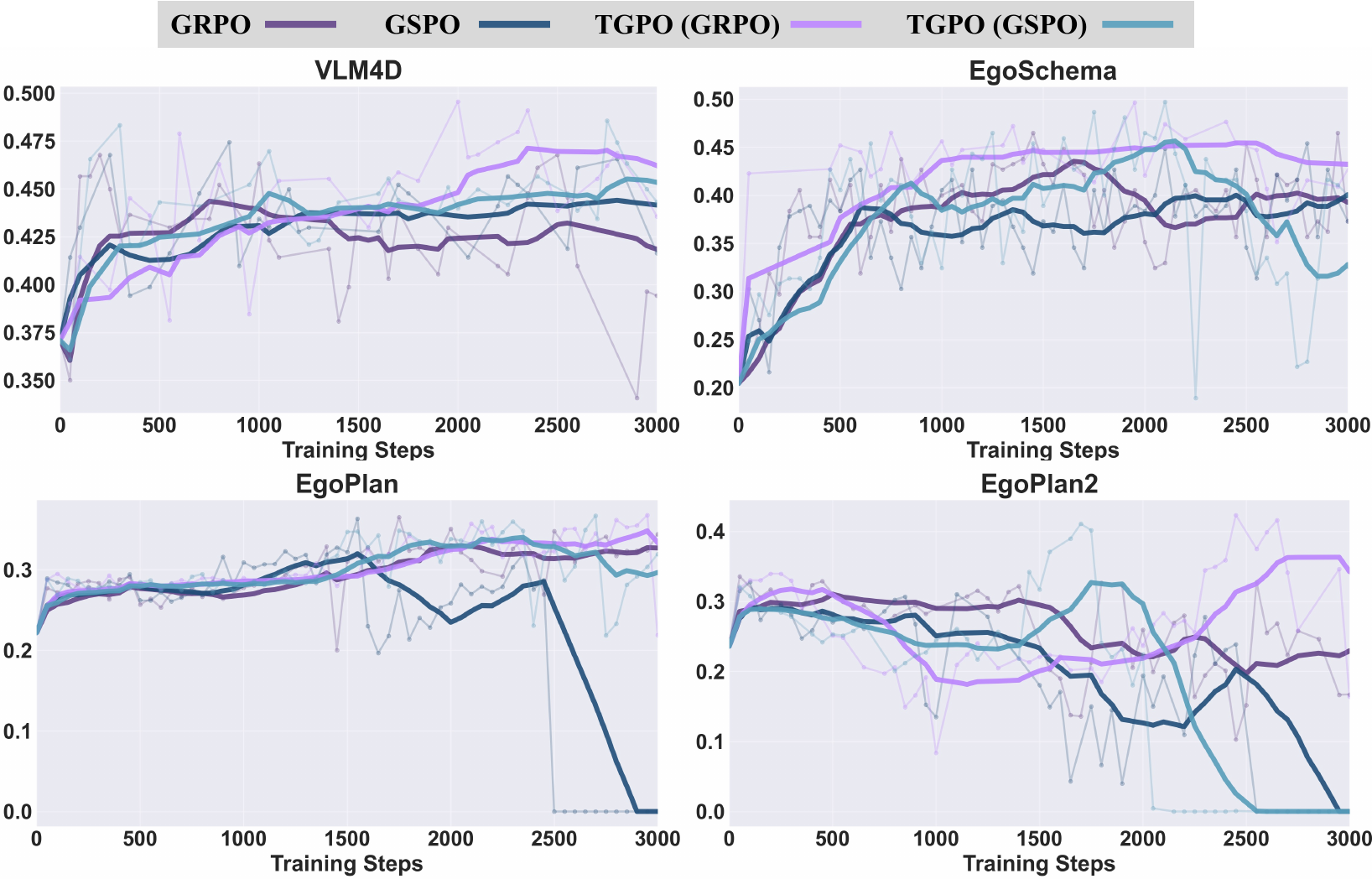}
  \caption{Test reward over the 3000 training steps. The reward curves are reported on four benchmarks across GRPO, GSPO, and our two \trpo{} variants.}
  \label{fig:training_dynamics}
\end{figure*}

We present the training dynamics (reward versus training steps) on four benchmarks in Figure~\ref{fig:training_dynamics}. Darker curves denote smoothed performance for improved readability, while lighter curves show raw reward trajectories. We exclude EgoTempo from this analysis, as its long video sequences substantially slow down training. As shown in the figure, both \trpo{} variants consistently achieve higher reward than their baseline counterparts throughout most of the training process across the evaluated benchmarks. In addition, \trpo{} exhibits faster early-stage reward growth and smoother training trajectories, particularly on VLM4D and EgoSchema, indicating improved learning efficiency and more stable optimization dynamics. Notably, GSPO shows a clear late-stage performance degradation on EgoPlan, whereas TGPO (GRPO) maintains a steady improvement trend. On the more challenging EgoPlan2 benchmark, we observe pronounced reward degradation for GSPO and GRPO, as well as for TGPO (GSPO), while TGPO (GRPO) largely avoids this failure mode and preserves stable reward accumulation throughout training.

To quantitatively assess learning efficiency, we report the Area Under the Curve (AUC) of reward over the first 3000 training steps in Table~\ref{tab:auc_across_datasets}. In reinforcement learning, a higher AUC reflects faster reward improvement and sustained performance throughout training, capturing overall sample efficiency rather than isolated peak performance. Notably, TGPO (GRPO) improves AUC over GRPO from 1159.03 to 1306.32 on EgoSchema (+12.7\%) and from 1265.13 to 1339.49 on VLM4D (+5.9\%). TGPO (GSPO) also yields consistent gains, improving AUC over GSPO from 1305.99 to 1333.75 on VLM4D (+2.1\%). Overall, these results demonstrate that \trpo{} enables more effective and stable reward accumulation over time, leading to faster convergence and improved training efficiency across benchmarks.

\subsection{System-Level Comparison}

\begin{table*}[h!]
\centering
\begin{minipage}[t]{0.4\linewidth}
\centering
\vspace{0pt}
\resizebox{\linewidth}{!}{
\begin{tabular}{l c}
\toprule
\rowcolor{gray!15}
\textbf{Model} & \textbf{EgoSchema$\uparrow$} \\
\midrule
Qwen2.5-VL (3B)        & 20.4 \\
LLaVA-Video (7B)       & 34.1 \\
LLaVA-Video (72B)      & 42.7 \\
Llava-OneVision (72B)  & 36.2 \\
EgoVLM GRPO (3B)*      & 46.5 \\
\midrule
{\small\textbf{Qwen2.5-VL 3B}} \\
\rowcolor{TgpoBlue!20}\textsc{+TGPO} & \textbf{49.7} \\
\bottomrule
\end{tabular}}
\caption{Performance comparison on EgoSchema.}
\label{tab:egoschema}
\end{minipage}
\hspace{0.03\linewidth}
\begin{minipage}[t]{0.4\linewidth}
\centering
\vspace{0pt}
\resizebox{\linewidth}{!}{
\begin{tabular}{l c}
\toprule
\rowcolor{gray!15}
\textbf{Model} & \textbf{EgoPlan 2$\uparrow$} \\
\midrule
\multicolumn{2}{l}{\small\textbf{Proprietary}} \\
GPT-4o                     & 32.6 \\
\midrule
\multicolumn{2}{l}{\small\textbf{Open-source MLLMs}} \\
LLaVA-Video (7B)           & 25.3 \\
LLaVA-NeXT-Video (7B)      & 23.3 \\
Video-LLaMA-2 (7B)         & 23.0 \\
EgoVLM GRPO (3B)*          & 37.1 \\
\midrule
{\small\textbf{Qwen2.5-VL 3B}} \\
\rowcolor{TgpoPurple!20}\textsc{+TGPO} & \textbf{42.3} \\
\bottomrule
\end{tabular}}
\caption{Performance comparison on EgoPlan 2.}
\label{tab:egoplan2}
\end{minipage}
\end{table*}

\begin{table*}[h!]
\centering
\begin{minipage}[t]{0.4\linewidth}
\centering
\vspace{0pt}
\resizebox{\linewidth}{!}{
\begin{tabular}{l c}
\toprule
\rowcolor{gray!15}
\textbf{Model} & \textbf{VLM4D$\uparrow$} \\
\midrule
\multicolumn{2}{l}{\small\textbf{Proprietary}} \\
Gemini2.5-Pro        & 64.6 \\
GPT-4o               & 55.5 \\
Claude-Sonnet-4      & 52.6 \\
Grok-2-Vision        & 48.8 \\
\midrule
\multicolumn{2}{l}{\small\textbf{Open-source MLLMs}} \\
Qwen2.5-VL (3B)      & 37.1 \\
Qwen2.5-VL (7B)      & 42.3 \\
InternVideo2 (8B)    & 35.6 \\
Llava-OneVision (7B) & 36.8 \\
EgoVLM GRPO (3B)*    & 46.8 \\
\midrule
{\small\textbf{Qwen2.5-VL 3B}} \\
\rowcolor{TgpoPurple!20}\textsc{+TGPO} & \textbf{49.6} \\
\bottomrule
\end{tabular}}
\caption{Performance comparison on VLM4D.}
\label{tab:vlm4d}
\end{minipage}
\hspace{0.03\linewidth}
\begin{minipage}[t]{0.4\linewidth}
\centering
\vspace{0pt}
\resizebox{\linewidth}{!}{
\begin{tabular}{l c}
\toprule
\rowcolor{gray!15}
\textbf{Model} & \textbf{EgoPlan$\uparrow$} \\
\midrule
\multicolumn{2}{l}{\small\textbf{Proprietary}} \\
Gemini1.5-Pro         & 32.8 \\
GPT-4o                & 32.8 \\
\midrule
\multicolumn{2}{l}{\small\textbf{Open-source MLLMs}} \\
Qwen2.5-VL (3B)       & 32.9 \\
Qwen2.5-VL (7B)       & 33.0 \\
LLaVA-Video (7B)      & 33.6 \\
EgoVLM SFT (3B)*      & 32.1 \\
EgoVLM Dr. GRPO (3B)  & 33.0 \\
EgoVLM GRPO (3B)*     & 36.5 \\
\midrule
{\small\textbf{Qwen2.5-VL 3B}} \\
\rowcolor{TgpoPurple!20}\textsc{+TGPO} & \textbf{36.8} \\
\bottomrule
\end{tabular}}
\caption{Performance comparison on EgoPlan.}
\label{tab:egoplan}
\end{minipage}
\end{table*}

Tables \ref{tab:egoschema}–\ref{tab:egotempo} report performance on five egocentric video question answering benchmarks that emphasize temporal understanding and long-horizon reasoning. Across all datasets, \trpo{} consistently achieves the strongest performance among open-source models, outperforming supervised fine-tuning and standard reinforcement learning baselines under comparable model scales.

On EgoSchema (Table \ref{tab:egoschema}), \trpo{} improves upon the strongest open-source baseline (EgoVLM GRPO, 3B) by a clear margin (49.7 vs. 46.5), and surpasses substantially larger models such as LLaVA-Video (72B). A similar trend is observed on EgoPlan and EgoPlan 2 (Tables \ref{tab:egoplan} and \ref{tab:egoplan2}), where \trpo{} consistently outperforms both supervised fine-tuning and reinforcement learning baselines initialized from the same backbone. On EgoPlan 2 in particular, TGPO achieves a large gain over EgoVLM GRPO (42.3 vs. 37.1), indicating that temporally calibrated optimization is especially beneficial in more challenging planning scenarios. Notably, \trpo{} also outperforms proprietary systems such as GPT-4o on EgoPlan 2, despite the latter’s significantly larger scale and access to private training data.

\begin{table}[h!]
\centering
\resizebox{0.8\linewidth}{!}{
\begin{tabular}{l|c}
\toprule
\rowcolor{gray!15}
\textbf{Model} 
& \textbf{EgoTempo$\uparrow$} \\
\midrule

\multicolumn{2}{l}{\small\textbf{Proprietary}} \\
\midrule
Gemini-Flash & 39.1 \\
GPT-4o & 40.1 \\
Claude-3.5-Sonnet & 13.1 \\

\midrule
\multicolumn{2}{l}{\small\textbf{Open-source MLLMs}} \\
\midrule
Qwen2-VL (7B) & 26.1 \\
Qwen2-VL (72B) & 28.4 \\
LLaVA-OneVision (7B) & 23.3 \\
LLaVA-OneVision (72B) & 26.5 \\
LLaVA-NeXT-Video (34B) & 15.7 \\
EgoVLM GRPO (3B)* & 40.8 \\

\midrule
\multicolumn{2}{l}{\small\textbf{Qwen2.5-VL 3B}} \\
\midrule
\rowcolor{TgpoPurple!20}
+ \textsc{TGPO} & \textbf{45.2} \\

\bottomrule
\end{tabular}
}
\caption{Performance comparison on EgoTempo benchmark.}
\label{tab:egotempo}
\end{table}

On VLM4D (Table \ref{tab:vlm4d}), which evaluates temporal understanding in translational and rotational motions and perspective awareness, \trpo{} again yields consistent improvements over strong open-source baselines, including Qwen2.5-VL (7B) and EgoVLM GRPO (3B). While large proprietary models such as Gemini-2.5-Pro still achieve the highest performance, \trpo{} narrows the gap substantially, outperforming several proprietary models and highlighting the effectiveness of targeted temporal reinforcement learning.

Finally, on EgoTempo (Table \ref{tab:egotempo}), \trpo{} achieves the best overall performance among all models, including proprietary systems. Compared to EgoVLM GRPO, \trpo{} yields a notable improvement (45.2 vs. 40.8), underscoring its advantage in tasks that require fine-grained temporal alignment rather than static visual recognition. This result further supports the hypothesis that explicitly optimizing temporal decision-making is critical for egocentric video understanding.

Overall, these results demonstrate that \trpo{} delivers consistent and robust gains across diverse egocentric benchmarks, and that temporally calibrated reinforcement learning can substantially reduce the performance gap between open-source and proprietary multimodal large language models without increasing model scale.

\section{Conclusion}

This work highlights the importance of explicitly modeling temporal structure in multimodal large language models for egocentric video understanding. By framing temporal awareness as a learnable and incentivized capability rather than an emergent property of post-training, we introduce \trpo{} as a principled reinforcement learning approach that directly targets causal and temporal reasoning. Our results demonstrate that contrastive temporal rewards and cold-start RL training can effectively guide MLLMs toward coherent, temporally grounded reasoning without reliance on supervised finetuning. We hope this perspective encourages future research to move beyond static visual reasoning and toward learning-based frameworks that better align multimodal models with the dynamic, causal nature of real-world perception.